\documentclass[11pt,a4paper]{article}
\usepackage[hyperref]{acl2019}
\pdfoutput=1
\usepackage{times}
\usepackage{latexsym}
\usepackage{amsmath}
\usepackage{url}
\usepackage{amssymb}
\usepackage{amsfonts}
\usepackage{graphicx}
\usepackage{tabularx}
\usepackage{multirow}
\usepackage{arydshln}
\usepackage{mathtools,nccmath}

\usepackage[utf8]{inputenc}

\aclfinalcopy 

\title{A neural joint model for  Vietnamese  word segmentation, POS tagging \\ and dependency parsing}

\author{Dat Quoc Nguyen$^{1,2}$  \\
$^{1}$The University of Melbourne, Australia \\
{\tt dqnguyen@unimelb.edu.au} \\
$^{2}$VinAI Research, Hanoi, Vietnam \\
{\tt v.datnq9@vinai.io}
}

\begin{document}
\maketitle

\begin{abstract}
We propose the {first} multi-task learning model for joint Vietnamese word segmentation, part-of-speech (POS) tagging and dependency parsing. In particular, our model extends the BIST graph-based dependency parser \citep{TACL885} with BiLSTM-CRF-based neural layers \citep{HuangXY15} for word segmentation and POS tagging. On Vietnamese benchmark  datasets, experimental  results  show that our joint  model obtains state-of-the-art or competitive performances.
\end{abstract}

\section{Introduction}

Dependency parsing \citep{Kubler2009} is extremely useful in many downstream applications such as relation extraction \citep{bunescu-mooney:2005:HLTEMNLP} and machine translation \citep{galley-manning:2009:ACLIJCNLP}. POS tags are essential features used  in  dependency parsing.  In real-world parsing, most parsers are used in a pipeline process with a  precursor POS tagging model for producing predicted POS tags.  In  English   where white space is a strong word boundary indicator, POS tagging is considered to be the first important step towards dependency  parsing \citep{ballesteros-etal-2015-improved}. 

\begin{figure}[!ht]
\centering
\includegraphics[width=7.5cm]{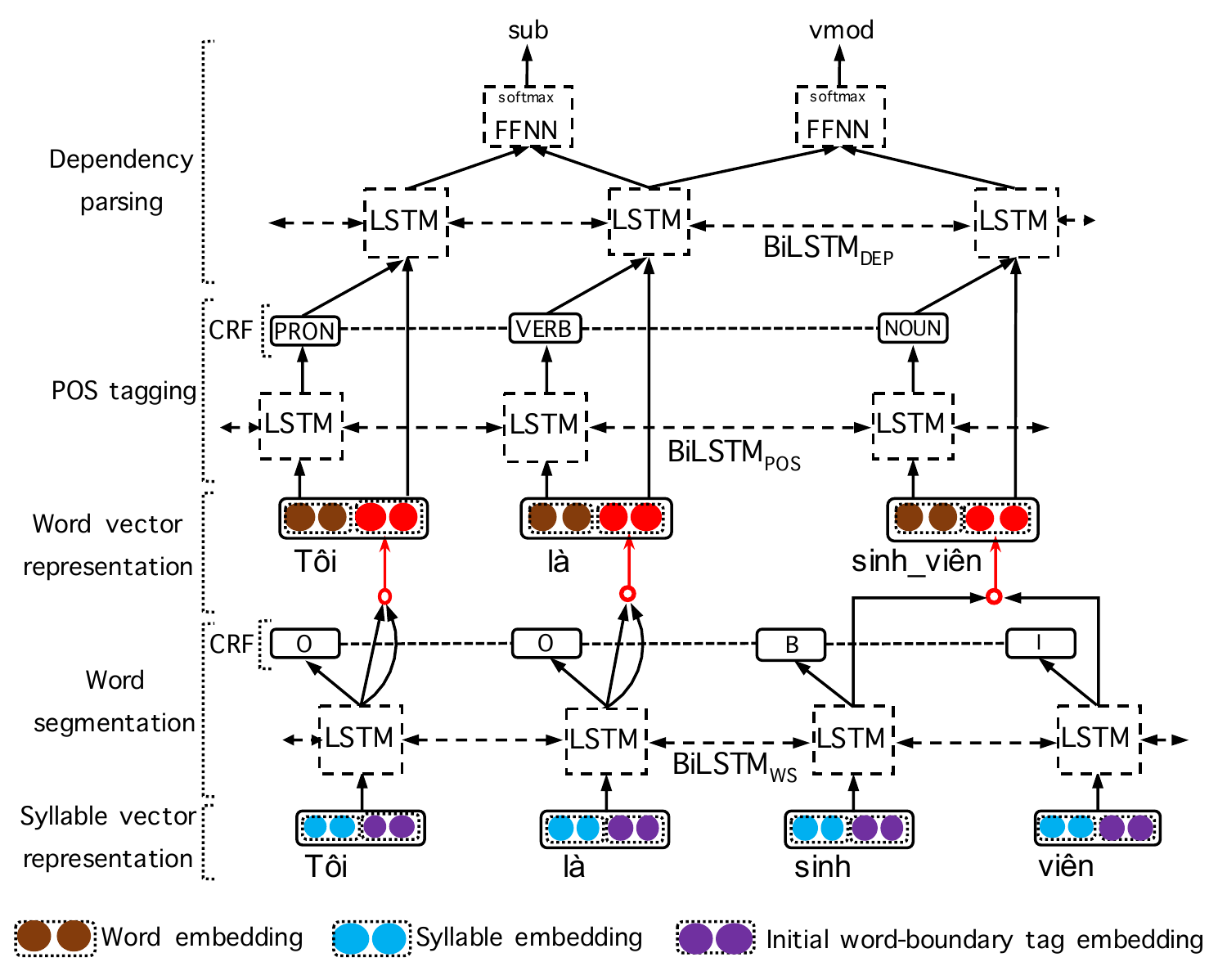}
{\small
\begin{tabular}{cllcl}
\hline 
ID & Form & POS & Head & DepRel \\
\hline
1 & Tôi\textsubscript{ I} & PRON& 2 & sub  \\
2 & là\textsubscript{ am} & VERB& 0 & root \\
3 & sinh\_viên\textsubscript{ student} & NOUN& 2 & vmod \\
\hline
\end{tabular}
}
\caption{Illustration of our joint model. Linear transformations are not shown   for simplification.}
\label{fig:architecture}
\end{figure}

Unlike English, for Vietnamese NLP, word segmentation is considered to be the key first step.  This is because when written,  white space is used in Vietnamese to separate syllables that constitute words, in addition to marking word boundaries \citep{nguyen-EtAl:2009:LAW-III}.   For example, a 4-syllable written text ``Tôi là sinh viên'' (I am student) forms 3 words ``Tôi\textsubscript{I} là\textsubscript{am} sinh\_viên\textsubscript{student}''.\footnote{About 85\% of  Vietnamese word types are composed of at least two syllables  and 80\%+ of syllable types are words by themselves \cite{DinhQuangThang2008}. For Vietnamese word segmentation,  white space is only used to separate word tokens while underscore is used to separate syllables inside a word.}  
When parsing  real-world Vietnamese text where gold word segmentation is not available, a pipeline process is defined that starts with a word segmenter to segment the text. The segmented text (e.g.\ ``Tôi là sinh\_viên'') is provided as the input to the POS tagger, which automatically generates POS-annotated text (e.g.\ ``Tôi/PRON là/VERB sinh\_viên/NOUN'') which is in turn fed to the parser. See Figure \ref{fig:architecture} for the final parsing output.  

However, Vietnamese word segmenters and POS taggers have a non-trivial error rate, thus leading to error propagation. 
A solution to these problems is to develop models for jointly learning  word segmentation, POS tagging and dependency parsing, such as those that have been actively explored for Chinese.    These include  traditional feature-based  models  \citep{hatoriACL2012,qianEMNLP-CoNLL,zhangP14,zhangNAACL} and neural models   \citep{kurita2017,LiZJZ18}. These models  construct \textit{transition}-based frameworks at character level.

In this paper, we present a new multi-task learning model for joint word segmentation, POS tagging and dependency parsing.  More specifically, our model can be viewed as an extension of the BIST \textit{graph}-based dependency parser \citep{TACL885}, that incorporates BiLSTM-CRF-based architectures \citep{HuangXY15} to predict the  segmentation and POS tags. To the best of our knowledge, our model is  the {first} one which is proposed to  jointly learn these three tasks for Vietnamese.  Experiments on Vietnamese benchmark datasets show that our   model  produces state-of-the-art or competitive results.

\section{Our proposed model}

As illustrated in Figure \ref{fig:architecture}, our  joint multi-task model can be
viewed as a hierarchical mixture of three components: word
segmentation, POS tagging and dependency parsing.  
In particular, our word segmentation component formalizes the  Vietnamese word segmentation  task as a sequence labeling problem, thus uses  a  BiLSTM-CRF architecture \citep{HuangXY15} to predict BIO word boundary tags from input syllables, resulting in a word-segmented  sequence. As for word segmentation, our POS tagging component also uses  a BiLSTM-CRF to predict POS tags from the sequence of segmented words. Based on the input segmented words and their  predicted POS tags, our dependency parsing component uses  a graph-based architecture  similarly to the one from  \citet{TACL885} to decode dependency arcs and labels. 

\paragraph{Syllable vector representation:}\ Given an input sentence $\mathcal{S}$ of $m$ syllables $s_1$, $s_2$, ..., $s_m$, we apply an initial word segmenter to produce {initial} BIO word-boundary tags  $b_1$, $b_2$, ..., $b_m$.  Following the state-of-the-art Vietnamese  word segmenter VnCoreNLP's RDRsegmenter \citep{NguyenNVDJ2018}, our initial word segmenter is  based on the lexicon-based longest matching strategy  \cite{Poowarawan}. We create a vector  $\mathbf{v}_i$  to represent each $i^{th}$ syllable in the input sentence $\mathcal{S}$ by concatenating its syllable embedding  $\mathbf{e}^{(\textsc{s})}_{s_i}$ and its  initial word-boundary tag embedding $\mathbf{e}^{(\textsc{b})}_{b_i}$: 

\setlength{\abovedisplayskip}{5pt}
\setlength{\belowdisplayskip}{5pt}
 \begin{equation}
\mathbf{v}_{i} =  \mathbf{e}^{(\textsc{s})}_{s_i} \circ \mathbf{e}^{(\textsc{b})}_{b_i}
\label{equation:syllableembed}
\end{equation}

\paragraph{Word segmentation (WSeg):}\ The WSeg component uses a BiLSTM ($\mathrm{BiLSTM}_{\textsc{ws}}$)  to learn a latent feature vector  representing the $i^{th}$ syllable from a sequence of vectors $\mathbf{v}_{1:m}$: 

\begin{equation}
\mathbf{r}_{i}^{(\textsc{ws})} =  \mathrm{BiLSTM}_{\textsc{ws}}(\mathbf{v}_{1:m}, i)
\label{equal:syllablembed}
\end{equation} 

The  WSeg component then uses a single-layer feed-forward  network ($\mathrm{FFNN}_{\textsc{ws}}$) to perform linear transformation over  each latent feature vector:

\begin{equation}
\mathbf{h}_{i}^{(\textsc{ws})} =  \mathrm{FFNN}_{\textsc{ws}}\big(\mathbf{r}_{i}^{(\textsc{ws})}\big)
\end{equation} 

 Next, the WSeg component feeds  output vectors $\mathbf{h}_{i}^{(\textsc{ws})}$  into a linear-chain CRF layer \cite{Lafferty:2001} for final BIO word-boundary tag prediction. A cross-entropy objective loss $\mathcal{L}_{\textsc{ws}}$ is computed during training, while the Viterbi algorithm is used for decoding.

\paragraph{Word vector representation:}\ Assume that we form $n$  words $w_1$, $w_2$, ..., $w_n$ based on $m$ syllables in the input sentence $\mathcal{S}$. Note that we use gold word segmentation when training, and use predicted  segmentation produced by the  WSeg component when decoding. We create a vector $\mathbf{x}_{j}$ to  represent each $j^{th}$ word  $w_j$ by concatenating its word embedding  $\mathbf{e}^{(\textsc{w})}_{w_j}$ and its syllable-level word embedding $\mathbf{e}^{(\textsc{sw})}_{w_j}$:

 \begin{equation}
\mathbf{x}_{j} =  \mathbf{e}^{(\textsc{w})}_{w_j} \circ \mathbf{e}^{(\textsc{sw})}_{w_j}
\label{eqution:xj}
\end{equation} 

Here, inspired by \citet{bohnet-etal-2018-morphosyntactic},  to obtain  $\mathbf{e}^{(\textsc{sw})}_{w_j}$, we combine sentence-level context sensitive syllable encodings (from  Equation \ref{equal:syllablembed}) and feed it into a FFNN ($\mathrm{FFNN}_{\textsc{sw}}$):

 \begin{equation}
\mathbf{e}^{(\textsc{sw})}_{w_j} =  \mathrm{FFNN}_{\textsc{sw}}\big( \mathbf{r}_{f(w_j)}^{(\textsc{ws})} \circ \mathbf{r}_{l(w_j)}^{(\textsc{ws})}  \big) 
\label{equation:swembed}
\end{equation} 

\noindent where $f(w_j)$ and $l(w_j)$ denote  indices of the first and last syllables of $w_j$ in  $\mathcal{S}$, respectively.

\paragraph{POS tagging:}\ The POS tagging component first feeds a sequence of vectors $\mathbf{x}_{1:n}$  into a BiLSTM ($\mathrm{BiLSTM}_{\textsc{pos}}$) to learn latent feature vectors representing input words, and passes each of these latent vectors as input to a FFNN ($\mathrm{FFNN}_{\textsc{pos}}$):

\begin{eqnarray}
\mathbf{r}_{j}^{(\textsc{pos})} &=&  \mathrm{BiLSTM}_{\textsc{pos}}(\mathbf{x}_{1:n}, j) \\
\mathbf{h}_{j}^{(\textsc{pos})} &=&  \mathrm{FFNN}_{\textsc{pos}}\big(\mathbf{r}_{j}^{(\textsc{pos})}\big)
\end{eqnarray}

Output vectors $\mathbf{h}_{j}^{(\textsc{pos})}$  are then fed  into a  CRF layer for POS tag prediction. A cross-entropy   loss $\mathcal{L}_{\textsc{pos}}$   is computed for POS tagging when training.

\paragraph{Dependency parsing:}\ Assume that the POS tagging component produces $p_1$, $p_2$, ..., $p_n$ as  predicted POS tags for the  input words $w_1$, $w_2$, ..., $w_n$, respectively. Each $j^{th}$  predicted POS tag $p_j$ is represented by an embedding $\mathbf{e}^{(\textsc{p})}_{p_j}$. We create  a sequence of vectors $\mathbf{z}_{1:n}$  as input for the dependency parsing component,  in which each $\mathbf{z}_j$  is resulted by concatenating the word vector representation $\mathbf{x}_j$ (from Equation \ref{eqution:xj}) and the corresponding POS tag embedding  $\mathbf{e}^{(\textsc{p})}_{p_j}$. The dependency parsing component uses a BiLSTM ($\mathrm{BiLSTM}_{\textsc{dep}}$) to learn latent feature representations from the input  $\mathbf{z}_{1:n}$:

 \begin{eqnarray}
\mathbf{z}_{j} &=&  \mathbf{x}_j  \circ \mathbf{e}^{(\textsc{p})}_{p_j} \label{equation:zj}  \\
\mathbf{r}_{j}^{(\textsc{dep})} &=&  \mathrm{BiLSTM}_{\textsc{dep}}(\mathbf{z}_{1:n}, j) \label{equation:rjdep}
\end{eqnarray}

Based on latent feature vectors $\mathbf{r}_{j}^{(\textsc{dep})}$, either a  transition-based or graph-based neural architecture can be applied for dependency parsing \citep{TACL885}. 

\citet{NguyenALTA2016} show that   in
both neural network-based and traditional feature-based categories, graph-based parsers perform better than  transition-based parsers for Vietnamese. Thus, our parsing component  is constructed similarly to the BIST graph-based dependency parser from \citet{TACL885}. A difference is that we  use FFNNs to split $\mathbf{r}_{j}^{(\textsc{dep})}$ into \textit{head} and \textit{dependent} representations: 

 \begin{eqnarray}
\mathbf{h}_{j}^{(\textsc{a-h})}  &=& \mathrm{FFNN}_{\text{Arc-Head}}\big(\mathbf{r}_{j}^{(\textsc{dep})}\big) \label{equation:ah}  \\ 
\mathbf{h}_{j}^{(\textsc{a-d})}  &=& \mathrm{FFNN}_{\text{Arc-Dep}}\big(\mathbf{r}_{j}^{(\textsc{dep})}\big)  \\ 
\mathbf{h}_{j}^{(\textsc{l-h})}  &=& \mathrm{FFNN}_{\text{Label-Head}}\big(\mathbf{r}_{j}^{(\textsc{dep})}\big)  \\ 
\mathbf{h}_{j}^{(\textsc{l-d})}  &=& \mathrm{FFNN}_{\text{Label-Dep}}\big(\mathbf{r}_{j}^{(\textsc{dep})}\big) \label{equation:ld} 
\end{eqnarray}

To score a potential dependency arc, we use a FFNN ($\mathrm{FFNN}_{\textsc{arc}}$) with a one-node output layer:

\begin{equation}
\text{score}(i,j) = \mathrm{FFNN}_{\textsc{arc}}\big( \mathbf{h}_{i}^{(\textsc{a-h})} \circ\ \mathbf{h}_{j}^{(\textsc{a-d})}  \big)
\end{equation}

\noindent Given  scores of word pairs, we  predict the highest scoring projective
parse tree by using the \citet{Eisner1996}
decoding algorithm. This unlabeled parsing model is trained with a margin-based hinge loss  $\mathcal{L}_{\textsc{arc}}$ \citep{TACL885}.  

To label predicted arcs, we use another FFNN ($\mathrm{FFNN}_{\textsc{label}}$) with  softmax output: 

\begin{equation}
\boldsymbol{v}_{(i,j)} = \mathrm{FFNN}_{\textsc{label}}\big( \mathbf{h}_{i}^{(\textsc{l-h})} \circ\ \mathbf{h}_{j}^{(\textsc{l-d})}  \big)
\end{equation}

Based on vectors $\boldsymbol{v}_{(i,j)}$, a cross entropy loss $\mathcal{L}_{\textsc{label}}$ for dependency label prediction  is computed when training, using the gold labeled  tree. 

\paragraph{Joint multi-task learning:}\ We train our model by summing  $\mathcal{L}_{\textsc{ws}}$, $\mathcal{L}_{\textsc{pos}}$, $\mathcal{L}_{\textsc{arc}}$ and $\mathcal{L}_{\textsc{label}}$ losses prior to computing gradients. Model parameters are  learned to minimize the sum of the losses.





\paragraph{Discussion:} 
Our model is inspired by stack propagation based methods \citep{zhang-weiss-2016-stack,hashimoto-etal-2017-joint} which are joint  models for POS tagging and dependency parsing. For dependency parsing, the Stack-propagation model \citep{zhang-weiss-2016-stack} uses a transition-based approach, and  the joint multi-task  model JMT \citep{hashimoto-etal-2017-joint} uses a head selection based approach which produces a probability distribution over possible heads for each word \citep{zhang-cheng-lapata:2017:EACLlong}, while our model uses a graph-based approach. 

Our model can be viewed as an extension of the joint POS tagging and dependency parsing model jPTDP-v2   \citep{nguyenverspoorK18},\footnote{On  the benchmark English PTB-WSJ corpus, jPTDP-v2  does better than Stack-propagation, while obtaining similar performance to JMT.} where we incorporate a BiLSTM-CRF for word boundary prediction. Other improvements to jPTDP-v2 include: (i) instead of using `local' single word-based character-level embeddings, we use `global' sentence-level context for learning word embeddings (see equations \ref{equal:syllablembed} and \ref{equation:swembed}), (ii) we  use a CRF layer for POS tagging instead of a softmax layer, and (iii) following \citet{DozatM17}, we employ \textit{head} and \textit{dependent} projection representations  (in Equations \ref{equation:ah}--\ref{equation:ld}) as  feature vectors for dependency parsing rather than  the top recurrent states (in Equation \ref{equation:rjdep}).

\section{Experimental setup}

\paragraph{Datasets:} We follow the setup used in the  Vietnamese NLP toolkit VnCoreNLP  \citep{vuN18}.  

For word segmentation and POS tagging, we use standard datasets from the Vietnamese Language and Speech Processing (VLSP) 2013  shared tasks.\footnote{\url{http://vlsp.org.vn/vlsp2013}}  To train the word segmentation layer, we use 75K manually word-segmented  sentences in which 70K sentences are used for training and 5K sentences are used for development. For POS tagging, we use 27,870 manually word-segmented and POS-annotated sentences in which 27K and 870 sentences are used for training and development, respectively. For both tasks, the test set consists of 2120 manually word-segmented and POS-annotated sentences. 

To train the dependency parsing layer, we use the benchmark Vietnamese dependency treebank VnDT ({v1.1}) of 10,197 sentences \citep{Nguyen2014NLDB}, and follow a standard split to use 1,020 sentences for test, 200 sentences for development and the remaining 8,977 sentences for training. 

\paragraph{Implementation:}\ We implement our model (namely, \textbf{jointWPD}) using  \textsc{DyNet} \cite{dynet}. We learn model parameters  using Adam  \cite{KingmaB14}, and run for 50 epochs.  We  compute the average of F$_1$ scores computed for word segmentation, POS tagging and (LAS) dependency parsing   after each training epoch. We choose the model with the highest average score over the development sets to apply to the test sets. See Appendix for implementation details.

\section{Main results}

\paragraph{End-to-end results:}\ Our scores on the test sets are presented in Table \ref{tab:results}.  We compare our scores with the VnCoreNLP toolkit \citep{vuN18} which produces  the previous highest reported results on the same test sets for the three tasks.    
Note that published scores of VnCoreNLP  for POS tagging and dependency parsing were reported using gold word segmentation, and its published scores for dependency parsing were reported using the previous VnDT v1.0.  
As the current released VnCoreNLP version is retrained using the VnDT v1.1 and we also use the same experimental setup, we thus  rerun  VnCoreNLP on the unsegmented test sentences and compute its scores.\footnote{See accuracy results w.r.t. the gold word segmentation in Table \ref{tab:addresults} in the  Appendix.} Our   jointWPD  obtains a slightly lower word segmentation score and a similar POS tagging score against VnCoreNLP. However, jointWPD  achieves  2.7\% absolute higher LAS and UAS  than VnCoreNLP.

\begin{table}[!t]
\centering
\resizebox{7.5cm}{!}{
\setlength{\tabcolsep}{0.25em}
\begin{tabular}{ll|llll}
\hline
\multicolumn{2}{c|}{{\textbf{Model}}}  & \textbf{WSeg} & \textbf{PTag} & \textbf{LAS} & \textbf{UAS} \\
\hline
\multirow{4}{*}{\rotatebox[origin=c]{90}{\footnotesize{Unsegmented}}}  & Our jointWPD & {97.81} & 94.05 & 71.50 &  77.23  \\
\cline{2-6}
& VnCoreNLP &   {97.90} & {94.06} & 68.84$^{**}$ & 74.52$^{**}$   \\
& jPTDP-v2 &  97.90  & 93.82$^*$  & 70.78$^{**}$ & 76.80$^{*}$  \\
& Biaffine &  97.90  & {94.06} & 72.59$^{**}$ & 78.54$^{**}$ \\
\hline 
\end{tabular}
}
\caption{F$_1$ scores (in \%) for word segmentation
(WSeg), POS tagging (PTag) and dependency parsing (LAS and UAS)  on  \emph{test} sets of {unsegmented} sentences. 
Scores    are computed on all tokens (including punctuation), employing the CoNLL 2017 shared task evaluation script \protect{\cite{zeman-EtAl}}. In all tables, $*$ and $**$ denote the statistically significant differences against jointWPD at p $\leq$ 0.05 and p $\leq$ 0.01, respectively. We compute sentence-level scores for each model and task, then use paired t-test to measure the significance level. }
\label{tab:results}
\end{table}

We also show  in Table \ref{tab:results} scores of the joint POS tagging and dependency parsing model jPTDP-v2  \citep{nguyenverspoorK18} and the  state-of-the-art  Biaffine dependency parser  \citep{DozatM17}. 
For Biaffine which requires automatically predicted POS tags, following \citet{vuN18}, we produce the predicted POS tags on the whole VnDT treebank by using VnCoreNLP. We train both  jPTDP-v2 and Biaffine with gold word segmentation.\footnote{We reimplement jPTDP-v2 such  that  its POS tagging layer makes use of the VLSP 2013 POS tagging training set of 27K sentences, and then perform hyper-parameter tuning. The original jPTDP-v2 implementation only uses gold POS tags  available in 8,977 training dependency trees, thus giving lower parsing performance than ours. For  Biaffine, we use its updated  version \citep{dozat-qi-manning:2017:K17-3} which won the CoNLL 2017 shared task on multilingual Universal Dependencies (UD) parsing from raw text \cite{zeman-EtAl}. Biaffine was also employed in all the top systems at the follow-up  CoNLL 2018 shared task  \cite{zeman-etal-2018-conll}.}   
For test, these models are fed with predicted word-segmented test sentences produced by VnCoreNLP. Our jointWPD performs  significantly better than jPTDP-v2 on both POS tagging and dependency parsing tasks. However, jointWPD obtains 1.1+\% lower LAS and UAS than Biaffine which  uses a  ``\textit{biaffine}'' 
attention mechanism  for predicting
dependency arcs and labels. We will  extend our  parsing component with the {biaffine} attention mechanism to investigate the benefit for our joint  model  in future work.

\begin{table}[!t]
\centering
\resizebox{7.5cm}{!}{
\setlength{\tabcolsep}{0.25em}
\begin{tabular}{l|llll}
\hline
Model & \textbf{WSeg} & \textbf{PTag} & \textbf{LAS} & \textbf{UAS} \\
\hline 
WS $\mapsto$ Pos $\mapsto$ Dep & 98.48$^{*}$ & 95.09$^{*}$ & 70.68$^{*}$ & 76.70$^{*}$ \\
\hline
Our jointWPD &  98.66  & 95.35 & 71.13 & 77.01 \\
\hdashline
\ \ (a) w/o Initial\textsubscript{BIO} & 98.25$^{**}$ & 95.01$^{*}$ & 70.34$^{**}$ & 76.36$^{**}$ \\
\ \ (b) w/o CRF\textsubscript{WSeg} & 98.32$^{**}$ & 95.06$^{*}$ &  70.48$^{**}$ & 76.47$^{**}$\\
\ \ (c) w/o CRF\textsubscript{PTag} & 98.65 & 95.14$^{*}$  & 71.00 & 76.94 \\
\ \ (d) w/o PTag & 98.63 & 95.10$^{*}$  & 69.78$^{**}$ & 76.03$^{**}$ \\
\hline 
\end{tabular}
}
\caption{F$_1$ scores on \emph{development} sets of unsegmented  sentences. (a): Without using {initial} word-boundary tag embedding, i.e., Equation \ref{equation:syllableembed} becomes $\mathbf{v}_{i} =  \mathbf{e}^{(\textsc{s})}_{s_i}$; (b):  Using a softmax layer for word-boundary tag prediction instead of a CRF layer; (c):  Using a softmax layer for POS tag prediction instead of a CRF layer; (d): Without using the POS tag embeddings for the parsing component, i.e. Equation \ref{equation:zj}  becomes $\mathbf{z}_{j} =  \mathbf{x}_j$.}
\label{tab:ablation}
\end{table}

\paragraph{Ablation analysis:}\ Table \ref{tab:ablation} shows performance of a {Pipeline} strategy WS $\mapsto$ Pos $\mapsto$ Dep where we treat our word segmentation, POS tagging and dependency parsing  components as independent networks, and train them separately. We find that  jointWPD does {significantly} better than the Pipeline strategy on all three tasks. 

Table \ref{tab:ablation} also presents ablation tests over 4 factors. When not using either initial word-boundary tag embeddings  or  the CRF layer for word-boundary tag prediction, all scores degrade by about 0.3+\% absolutely.  The 2 remaining factors, including (c) using a softmax classifier for POS tag prediction rather than a CRF layer and (d) removing POS tag embeddings, do not effect the word segmentation score. Both factors notably decrease the POS tagging score.  Factor (c) slightly decreases LAS and UAS parsing scores. Factor (d) degrades the parsing scores by about 1.0+\%, clearly showing the usefulness of POS tag information for the dependency  parsing task. 

\section{Related work}

\citet{NguyenNVDJ2018}    propose a transformation rule-based learning model RDRsegmenter for Vietnamese word segmentation,  which obtains the highest performance  to date. 
\citet{NguyenALTA2017} briefly review word segmentation and POS tagging approaches for Vietnamese. In addition, \citet{NguyenALTA2017} also present an empirical comparison between state-of-the-art feature- and neural network-based models for Vietnamese POS tagging, and show that a conventional 
feature-based model performs better than neural network-based models. In particular,  on the VLSP 2013 POS tagging dataset, MarMoT \citep{mueller-etal-2013-efficient} obtains better accuracy than BiLSTM-CRF-based models with LSTM- and CNN-based character level word embeddings \citep{lample-EtAl:2016:N16-1,ma-hovy:2016:P16-1}.  \citet{vuN18}  incorporate RDRsegmenter and MarMoT  as the word segmentation and POS tagging components of VnCoreNLP, respectively. 

\citet{Thi2013} propose  a conversion method to automatically convert the manually built Vietnamese  constituency treebank \citep{nguyen-EtAl:2009:LAW-III} into  a dependency treebank. 
However,  \citet{Thi2013} do not clarify how dependency labels are inferred; also, they ignore syntactic information encoded in grammatical function tags, and  unable to deal with coordination and empty category cases.\footnote{\citet{Thi2013} reformed their dependency  treebank with the UD annotation scheme to create a Vietnamese UD treebank in 2017. Note that the CoNLL 2017 \& 2018 multilingual parsing shared tasks also provided F$_1$ scores for word segmentation, POS tagging and dependency parsing on this Vietnamese UD treebank. However, this UD treebank is small (containing  about 1,400 training sentences), thus it might not be ideal to draw a reliable conclusion.}  
\citet{Nguyen2014NLDB} later present a new conversion method to tackle all those issues, producing the high quality dependency treebank VnDT which is then    widely used in Vietnamese dependency parsing research  \citep{7371762,7758049,NguyenALTA2016,8573397,vuN18}.  
Recently, \citet{NGUYEN18.69} manually builds another Vietnamese dependency treebank---BKTreebank---consisting of about 7K sentences based on the Stanford Dependencies annotation scheme \citep{Marneffe2008}. 

\section{Conclusions and future work}

In this paper, we have presented the first multi-task learning model for joint word segmentation,
POS tagging and dependency parsing in Vietnamese. Experiments on Vietnamese  benchmark  datasets show that our joint multi-task model obtains results competitive with the state-of-the-art. 

\citet{che-etal-2018-towards} 
show that deep contextualized word representations \citep{elmo,bert} help  improve the  parsing performance. We will evaluate effects of the contextualized representations to our joint model.  
A Vietnamese syllable is analogous to a character in other languages such as Chinese and Japanese. Thus we will also  evaluate the application of our  model to those languages in future work.

\section*{Acknowledgments} 
I would like to thank Karin Verspoor and   Vu Cong Duy Hoang  
 as well as the anonymous reviewers  
for their feedback. 

\vspace{-10pt}

\bibliography{refs}

\begin{thebibliography}{46}
\expandafter\ifx\csname natexlab\endcsname\relax\def\natexlab#1{#1}\fi

\bibitem[{Ballesteros et~al.(2015)Ballesteros, Dyer, and
  Smith}]{ballesteros-etal-2015-improved}
Miguel Ballesteros, Chris Dyer, and Noah~A. Smith. 2015.
\newblock Improved transition-based parsing by modeling characters instead of
  words with {LSTM}s.
\newblock In \emph{Proceedings of EMNLP}, pages 349--359.

\bibitem[{Bohnet et~al.(2018)Bohnet, McDonald, Sim{\~o}es, Andor, Pitler, and
  Maynez}]{bohnet-etal-2018-morphosyntactic}
Bernd Bohnet, Ryan McDonald, Gon{\c{c}}alo Sim{\~o}es, Daniel Andor, Emily
  Pitler, and Joshua Maynez. 2018.
\newblock {Morphosyntactic Tagging with a Meta-{B}i{LSTM} Model over Context
  Sensitive Token Encodings}.
\newblock In \emph{Proceedings of ACL}, pages 2642--2652.

\bibitem[{Bunescu and Mooney(2005)}]{bunescu-mooney:2005:HLTEMNLP}
Razvan Bunescu and Raymond Mooney. 2005.
\newblock {A Shortest Path Dependency Kernel for Relation Extraction}.
\newblock In \emph{Proceedings of HLT-EMNLP}, pages 724--731.

\bibitem[{Che et~al.(2018)Che, Liu, Wang, Zheng, and
  Liu}]{che-etal-2018-towards}
Wanxiang Che, Yijia Liu, Yuxuan Wang, Bo~Zheng, and Ting Liu. 2018.
\newblock Towards better {UD} parsing: Deep contextualized word embeddings,
  ensemble, and treebank concatenation.
\newblock In \emph{Proceedings of the {C}o{NLL} 2018 Shared Task}, pages
  55--64.

\bibitem[{Devlin et~al.(2019)Devlin, Chang, Lee, and Toutanova}]{bert}
Jacob Devlin, Ming{-}Wei Chang, Kenton Lee, and Kristina Toutanova. 2019.
\newblock {BERT: Pre-training of Deep Bidirectional Transformers for Language
  Understanding}.
\newblock In \emph{Proceedings of NAACL-HLT}.

\bibitem[{Dozat and Manning(2017)}]{DozatM17}
Timothy Dozat and Christopher~D. Manning. 2017.
\newblock {Deep Biaffine Attention for Neural Dependency Parsing}.
\newblock In \emph{Proceedings of ICLR}.

\bibitem[{Dozat et~al.(2017)Dozat, Qi, and
  Manning}]{dozat-qi-manning:2017:K17-3}
Timothy Dozat, Peng Qi, and Christopher~D. Manning. 2017.
\newblock {Stanford's Graph-based Neural Dependency Parser at the CoNLL 2017
  Shared Task}.
\newblock In \emph{Proceedings of the CoNLL 2017 Shared Task}, pages 20--30.

\bibitem[{Eisner(1996)}]{Eisner1996}
Jason~M. Eisner. 1996.
\newblock {Three New Probabilistic Models for Dependency Parsing: An
  Exploration}.
\newblock In \emph{Proceedings of COLING}, pages 340--345.

\bibitem[{Galley and Manning(2009)}]{galley-manning:2009:ACLIJCNLP}
Michel Galley and Christopher~D. Manning. 2009.
\newblock {Quadratic-Time Dependency Parsing for Machine Translation}.
\newblock In \emph{Proceedings of ACL-IJCNLP}, pages 773--781.

\bibitem[{Hashimoto et~al.(2017)Hashimoto, Xiong, Tsuruoka, and
  Socher}]{hashimoto-etal-2017-joint}
Kazuma Hashimoto, Caiming Xiong, Yoshimasa Tsuruoka, and Richard Socher. 2017.
\newblock A joint many-task model: Growing a neural network for multiple {NLP}
  tasks.
\newblock In \emph{Proceedings of EMNLP}, pages 1923--1933.

\bibitem[{Hatori et~al.(2012)Hatori, Matsuzaki, Miyao, and
  Tsujii}]{hatoriACL2012}
Jun Hatori, Takuya Matsuzaki, Yusuke Miyao, and Jun'ichi Tsujii. 2012.
\newblock {Incremental Joint Approach to Word Segmentation, POS Tagging, and
  Dependency Parsing in Chinese}.
\newblock In \emph{Proceedings of ACL}, pages 1045--1053.

\bibitem[{Huang et~al.(2015)Huang, Xu, and Yu}]{HuangXY15}
Zhiheng Huang, Wei Xu, and Kai Yu. 2015.
\newblock {Bidirectional {LSTM-CRF} Models for Sequence Tagging}.
\newblock \emph{arXiv preprint}, arXiv:1508.01991.

\bibitem[{Kingma and Ba(2014)}]{KingmaB14}
Diederik~P. Kingma and Jimmy Ba. 2014.
\newblock {Adam: {A} Method for Stochastic Optimization}.
\newblock \emph{arXiv preprint}, arXiv:1412.6980.

\bibitem[{Kiperwasser and Goldberg(2016)}]{TACL885}
Eliyahu Kiperwasser and Yoav Goldberg. 2016.
\newblock {Simple and Accurate Dependency Parsing Using Bidirectional LSTM
  Feature Representations}.
\newblock \emph{Transactions of ACL}, 4:313--327.

\bibitem[{K\"{u}bler et~al.(2009)K\"{u}bler, McDonald, and Nivre}]{Kubler2009}
Sandra K\"{u}bler, Ryan McDonald, and Joakim Nivre. 2009.
\newblock \emph{{Dependency Parsing}}.
\newblock Synthesis Lectures on Human Language Technologies, Morgan \& cLaypool
  publishers.

\bibitem[{Kurita et~al.(2017)Kurita, Kawahara, and Kurohashi}]{kurita2017}
Shuhei Kurita, Daisuke Kawahara, and Sadao Kurohashi. 2017.
\newblock {Neural Joint Model for Transition-based Chinese Syntactic Analysis}.
\newblock In \emph{Proceedings of ACL}, pages 1204--1214.

\bibitem[{Lafferty et~al.(2001)Lafferty, McCallum, and Pereira}]{Lafferty:2001}
John~D. Lafferty, Andrew McCallum, and Fernando C.~N. Pereira. 2001.
\newblock {Conditional Random Fields: Probabilistic Models for Segmenting and
  Labeling Sequence Data}.
\newblock In \emph{Proceedings of ICML}, pages 282--289.

\bibitem[{Lample et~al.(2016)Lample, Ballesteros, Subramanian, Kawakami, and
  Dyer}]{lample-EtAl:2016:N16-1}
Guillaume Lample, Miguel Ballesteros, Sandeep Subramanian, Kazuya Kawakami, and
  Chris Dyer. 2016.
\newblock {Neural Architectures for Named Entity Recognition}.
\newblock In \emph{Proceedings of NAACL-HLT}, pages 260--270.

\bibitem[{Li et~al.(2018)Li, Zhang, Ju, and Zhao}]{LiZJZ18}
Haonan Li, Zhisong Zhang, Yuqi Ju, and Hai Zhao. 2018.
\newblock {Neural Character-level Dependency Parsing for Chinese}.
\newblock In \emph{Proceedings of AAAI}, pages 5205--5212.

\bibitem[{Ma and Hovy(2016)}]{ma-hovy:2016:P16-1}
Xuezhe Ma and Eduard Hovy. 2016.
\newblock {End-to-end Sequence Labeling via Bi-directional LSTM-CNNs-CRF}.
\newblock In \emph{Proceedings of ACL}, pages 1064--1074.

\bibitem[{Marneffe and Manning(2008)}]{Marneffe2008}
Marie-catherine~De Marneffe and Christopher~D. Manning. 2008.
\newblock {The Stanford typed dependencies representation}.
\newblock In \emph{Proceedings of the Coling 2008 workshop on Cross-Framework
  and Cross-Domain Parser Evaluation}, pages 1--8.

\bibitem[{Mueller et~al.(2013)Mueller, Schmid, and
  Sch{\"u}tze}]{mueller-etal-2013-efficient}
Thomas Mueller, Helmut Schmid, and Hinrich Sch{\"u}tze. 2013.
\newblock Efficient higher-order {CRF}s for morphological tagging.
\newblock In \emph{Proceedings of EMNLP}, pages 322--332.

\bibitem[{Neubig et~al.(2017)Neubig, Dyer, Goldberg et~al.}]{dynet}
Graham Neubig, Chris Dyer, Yoav Goldberg, et~al. 2017.
\newblock {DyNet: The Dynamic Neural Network Toolkit}.
\newblock \emph{arXiv preprint}, arXiv:1701.03980.

\bibitem[{Nguyen et~al.(2018{\natexlab{a}})Nguyen, Nguyen, and
  Nguyen}]{8573397}
Binh~Duc Nguyen, Kiet~Van Nguyen, and Ngan Luu-Thuy Nguyen. 2018{\natexlab{a}}.
\newblock {LSTM Easy-first Dependency Parsing with Pre-trained Word Embeddings
  and Character-level Word Embeddings in Vietnamese}.
\newblock In \emph{Proceedings of KSE}, pages 187--192.

\bibitem[{Nguyen et~al.(2016)Nguyen, Dras, and Johnson}]{NguyenALTA2016}
Dat~Quoc Nguyen, Mark Dras, and Mark Johnson. 2016.
\newblock {An empirical study for Vietnamese dependency parsing}.
\newblock In \emph{Proceedings of ALTA}, pages 143--149.

\bibitem[{Nguyen et~al.(2014)Nguyen, Nguyen, Pham, Nguyen, and
  Nguyen}]{Nguyen2014NLDB}
Dat~Quoc Nguyen, Dai~Quoc Nguyen, Son~Bao Pham, Phuong-Thai Nguyen, and Minh~Le
  Nguyen. 2014.
\newblock {From Treebank Conversion to Automatic Dependency Parsing for
  Vietnamese}.
\newblock In \emph{{Proceedings of NLDB}}, pages 196--207.

\bibitem[{Nguyen et~al.(2018{\natexlab{b}})Nguyen, Nguyen, Vu, Dras, and
  Johnson}]{NguyenNVDJ2018}
Dat~Quoc Nguyen, Dai~Quoc Nguyen, Thanh Vu, Mark Dras, and Mark Johnson.
  2018{\natexlab{b}}.
\newblock {A Fast and Accurate Vietnamese Word Segmenter}.
\newblock In \emph{Proceedings of LREC}, pages 2582--2587.

\bibitem[{Nguyen and Verspoor(2018)}]{nguyenverspoorK18}
Dat~Quoc Nguyen and Karin Verspoor. 2018.
\newblock {An Improved Neural Network Model for Joint {POS} Tagging and
  Dependency Parsing}.
\newblock In \emph{Proceedings of the {CoNLL} 2018 Shared Task}, pages 81--91.

\bibitem[{Nguyen et~al.(2017)Nguyen, Vu, Nguyen, Dras, and
  Johnson}]{NguyenALTA2017}
Dat~Quoc Nguyen, Thanh Vu, Dai~Quoc Nguyen, Mark Dras, and Mark Johnson. 2017.
\newblock {From Word Segmentation to POS Tagging for Vietnamese}.
\newblock In \emph{Proceedings of ALTA}, pages 108--113.

\bibitem[{Nguyen(2018)}]{NGUYEN18.69}
Kiem-Hieu Nguyen. 2018.
\newblock {BKTreebank: Building a Vietnamese Dependency Treebank}.
\newblock In \emph{Proceedings LREC}, pages 2164--2168.

\bibitem[{Nguyen and Nguyen(2015)}]{7371762}
Kiet~Van Nguyen and Ngan Luu-Thuy Nguyen. 2015.
\newblock {Error Analysis for Vietnamese Dependency Parsing}.
\newblock In \emph{Proceedings of KSE}, pages 79--84.

\bibitem[{Nguyen and Nguyen(2016)}]{7758049}
Kiet~Van Nguyen and Ngan Luu-Thuy Nguyen. 2016.
\newblock {Vietnamese transition-based dependency parsing with supertag
  features}.
\newblock In \emph{Proceedings of KSE}, pages 175--180.

\bibitem[{Nguyen et~al.(2009)Nguyen, Vu, Nguyen, Nguyen, and
  Le}]{nguyen-EtAl:2009:LAW-III}
Phuong~Thai Nguyen, Xuan~Luong Vu, Thi Minh~Huyen Nguyen, Van~Hiep Nguyen, and
  Hong~Phuong Le. 2009.
\newblock {Building a Large Syntactically-Annotated Corpus of Vietnamese}.
\newblock In \emph{Proceedings of LAW}, pages 182--185.

\bibitem[{Peters et~al.(2018)Peters, Neumann, Iyyer, Gardner, Clark, Lee, and
  Zettlemoyer}]{elmo}
Matthew Peters, Mark Neumann, Mohit Iyyer, Matt Gardner, Christopher Clark,
  Kenton Lee, and Luke Zettlemoyer. 2018.
\newblock Deep contextualized word representations.
\newblock In \emph{Proceedings of NAACL-HLT}, pages 2227--2237.

\bibitem[{Poowarawan(1986)}]{Poowarawan}
Yuen Poowarawan. 1986.
\newblock {Dictionary-based Thai Syllable Separation}.
\newblock In \emph{Proceedings of the Ninth Electronics Engineering
  Conference}, pages 409--418.

\bibitem[{Qian and Liu(2012)}]{qianEMNLP-CoNLL}
Xian Qian and Yang Liu. 2012.
\newblock {Joint Chinese Word Segmentation, POS Tagging and Parsing}.
\newblock In \emph{Proceedings of EMNLP-CoNLL}, pages 501--511.

\bibitem[{Srivastava et~al.(2014)Srivastava, Hinton, Krizhevsky, Sutskever, and
  Salakhutdinov}]{JMLR:v15:srivastava14a}
Nitish Srivastava, Geoffrey Hinton, Alex Krizhevsky, Ilya Sutskever, and Ruslan
  Salakhutdinov. 2014.
\newblock {Dropout: A Simple Way to Prevent Neural Networks from Overfitting}.
\newblock \emph{Journal of Machine Learning Research}, 15:1929--1958.

\bibitem[{Thang et~al.(2008)Thang, Phuong, Huyen, Tu, Rossignol, and
  Luong}]{DinhQuangThang2008}
Dinh~Quang Thang, Le~Hong Phuong, Nguyen Thi~Minh Huyen, Nguyen~Cam Tu, Mathias
  Rossignol, and Vu~Xuan Luong. 2008.
\newblock {Word segmentation of Vietnamese texts: a comparison of approaches}.
\newblock In \emph{Proceedings of LREC}, pages 1933--1936.

\bibitem[{Thi et~al.(2013)Thi, My, Viet, Minh, and Hong}]{Thi2013}
Luong~Nguyen Thi, Linh~Ha My, Hung~Nguyen Viet, Huyen Nguyen~Thi Minh, and
  Phuong~Le Hong. 2013.
\newblock {Building a treebank for Vietnamese dependency parsing}.
\newblock In \emph{Proceedings of RIVF}, pages 147--151.

\bibitem[{Vu et~al.(2018)Vu, Nguyen, Nguyen, Dras, and Johnson}]{vuN18}
Thanh Vu, Dat~Quoc Nguyen, Dai~Quoc Nguyen, Mark Dras, and Mark Johnson. 2018.
\newblock {VnCoreNLP: A Vietnamese Natural Language Processing Toolkit}.
\newblock In \emph{Proceedings of NAACL: Demonstrations}, pages 56--60.

\bibitem[{Zeman et~al.(2017)}]{zeman-EtAl}
Daniel Zeman et~al. 2017.
\newblock {CoNLL 2017 Shared Task: Multilingual Parsing from Raw Text to
  Universal Dependencies}.
\newblock In \emph{Proceedings of the CoNLL 2017 Shared Task}, pages 1--19.

\bibitem[{Zeman et~al.(2018)}]{zeman-etal-2018-conll}
Daniel Zeman et~al. 2018.
\newblock {C}o{NLL} 2018 shared task: Multilingual parsing from raw text to
  universal dependencies.
\newblock In \emph{Proceedings of the {C}o{NLL} 2018 Shared Task}, pages 1--21.

\bibitem[{Zhang et~al.(2014)Zhang, Zhang, Che, and Liu}]{zhangP14}
Meishan Zhang, Yue Zhang, Wanxiang Che, and Ting Liu. 2014.
\newblock {Character-Level Chinese Dependency Parsing}.
\newblock In \emph{Proceedings of ACL}, pages 1326--1336.

\bibitem[{Zhang et~al.(2017)Zhang, Cheng, and
  Lapata}]{zhang-cheng-lapata:2017:EACLlong}
Xingxing Zhang, Jianpeng Cheng, and Mirella Lapata. 2017.
\newblock {Dependency Parsing as Head Selection}.
\newblock In \emph{Proceedings of EACL}, pages 665--676.

\bibitem[{Zhang et~al.(2015)Zhang, Li, Barzilay, and Darwish}]{zhangNAACL}
Yuan Zhang, Chengtao Li, Regina Barzilay, and Kareem Darwish. 2015.
\newblock {Randomized Greedy Inference for Joint Segmentation, POS Tagging and
  Dependency Parsing}.
\newblock In \emph{Proceedings of NAACL-HLT}, pages 42--52.

\bibitem[{Zhang and Weiss(2016)}]{zhang-weiss-2016-stack}
Yuan Zhang and David Weiss. 2016.
\newblock Stack-propagation: Improved representation learning for syntax.
\newblock In \emph{Proceedings of ACL}, pages 1557--1566.

\end{thebibliography}
\bibliographystyle{acl_natbib}

\appendix
\section*{Appendix}\label{appendix}

\paragraph{Implementation details:}\  
 When training, each task component is fed with the corresponding task-associated sentences. The dependency parsing training set is smallest in size (consisting of 8,977 sentences), thus  for each training epoch,  we sample the same number of sentences from the word segmentation and POS tagging training sets.  
We train our model with  a  fixed random  seed and without mini-batches. Dropout \cite{JMLR:v15:srivastava14a} is applied with a 67\% keep probability to the inputs of  BiLSTMs and FFNNs. Following  \citet{TACL885},  we also use \textit{word dropout} to learn  embeddings for unknown syllables/words: we replace each syllable/word  token $s/w$ appearing $\#(s/w)$ times with a ``unk'' symbol with probability $\mathsf{p}_{unk}(s/w) = \frac{0.25}{0.25 + \#(s/w)}$. 

We initialize syllable and word embeddings with 100-dimensional pre-trained Word2Vec vectors  as used in \citet{NguyenALTA2017}, while the initial word-boundary and POS tag embeddings are randomly initialized. All these embeddings are then updated when  training.  The sizes of the output layers of $\mathrm{FFNN}_{\textsc{ws}}$, $\mathrm{FFNN}_{\textsc{pos}}$ and $\mathrm{FFNN}_{\textsc{label}}$ are the number of BIO word-boundary tags (i.e.\ 3), the number of POS tags and the number of dependency relation types, respectively.  We perform a minimal grid search of hyper-parameters, resulting in the size of the initial word-boundary tag embeddings at 25, the POS tag embedding size of 100,  the size of the output layers of remaining FFNNs at 100,  the number of BiLSTM layers at 2 and the size of LSTM hidden states  in each layer  at 128.

\paragraph{Additional results:}\ Table \ref{tab:addresults} presents POS tagging and parsing accuracies  w.r.t. gold word segmentation.  In this case, for our jointWPD, we feed the POS tagging and parsing components with gold  word-segmented  sentences when decoding. 

\begin{table}[!ht]
\centering
\resizebox{7.5cm}{!}{
\setlength{\tabcolsep}{0.25em}
\begin{tabular}{ll|llll}
\hline
\multicolumn{2}{c|}{{\textbf{Model}}}  & \textbf{WSeg} & \textbf{PTag} & \textbf{LAS} & \textbf{UAS} \\
\hline 
\multirow{4}{*}{\rotatebox[origin=c]{90}{\footnotesize{Gold segment.}}}  & Our jointWPD & 100.0 & 95.97 & 73.90& 80.12 \\
\cline{2-6}
 & VnCoreNLP & 100.0 & 95.88 & 71.38$^{**}$ & 77.35$^{**}$\\
& jPTDP-v2 & 100.0 &  95.70$^{*}$ & 73.12$^{**}$ & 79.63$^{*}$ \\
& Biaffine & 100.0 & 95.88 & 74.99$^{**}$ & 81.19$^{**}$ \\
\hline
\end{tabular}
}
\caption{POS tagging, LAS and UAS accuracy scores  on the test sets w.r.t. {gold word-segmented}  sentences. These scores    are computed on all tokens (including punctuation). Recall that the LAS and UAS accuracies are computed on the VnDT v1.1 test set w.r.t. the  automatically predicted POS tags.}
\label{tab:addresults}
\end{table}

\end{document}